\documentclass[11pt,a4paper]{article}
\usepackage[hyperref]{emnlp2018}
\usepackage{times}
\usepackage{latexsym}
\usepackage{lipsum}
\usepackage{mathtools}
\usepackage{subcaption}
\usepackage{amssymb}
\usepackage{graphicx}
\usepackage{url}
\usepackage{array}
\usepackage[font={footnotesize}]{caption}

\aclfinalcopy % Uncomment this line for the final submission

\title{Compare, Compress and Propagate: Enhancing Neural Architectures with Alignment Factorization for Natural Language Inference}

\author{Yi Tay$^\dagger$, Luu Anh Tuan$^\ast$, Siu Cheung Hui$^\phi$ \\
  $^\dagger$$^\phi$Nanyang Technological University, Singapore \\
  $^\ast$Institute for Infocomm Research, A*Star Singapore \\
  {\tt ytay017@e.ntu.edu.sg$^\dagger$, at.luu@i2r.a-star.edu.sg$^\ast$} \\
  {\tt asschui@ntu.edu.sg$^\phi$} \\
  }

\date{}

\begin{document}
\maketitle
\begin{abstract}
This paper presents a new deep learning architecture for Natural Language Inference (NLI). Firstly, we introduce a new architecture where alignment pairs are compared, compressed and then propagated to upper layers for enhanced representation learning. Secondly, we adopt factorization layers for efficient and expressive compression of alignment vectors into scalar features, which are then used to augment the base word representations. The design of our approach is aimed to be conceptually simple, compact and yet powerful. We conduct experiments on three popular benchmarks, SNLI, MultiNLI and SciTail, achieving competitive performance on all. A lightweight parameterization of our model also enjoys a $\approx 3$ times reduction in parameter size compared to the existing state-of-the-art models, e.g., ESIM and DIIN, while maintaining competitive performance. Additionally, visual analysis shows that our propagated features are highly interpretable.
 \end{abstract}

 \section{Introduction}
Natural Language Inference (NLI) is a pivotal and fundamental task in language understanding and artificial intelligence. More concretely, given a premise and hypothesis, NLI aims to detect whether the latter \textit{entails} or \textit{contradicts} the former. As such, NLI is also commonly known as Recognizing Textual Entailment (RTE). NLI is known to be a significantly challenging task for machines whose success often depends on a wide repertoire of reasoning techniques.

In recent years, we observe a steep improvement in NLI systems, largely contributed by the release of the largest publicly available corpus for NLI - the Stanford Natural Language Inference (SNLI) corpus \cite{DBLP:conf/emnlp/BowmanAPM15} which comprises $570K$ hand labeled sentence pairs. This has improved the feasibility of training complex neural models, given the fact that neural models often require a relatively large amount of training data.

Highly competitive neural models for NLI are mostly based on soft-attention alignments, popularized by \cite{DBLP:conf/emnlp/ParikhT0U16,rocktaschel2015reasoning}. The key idea is to learn an alignment of sub-phrases in both sentences and learn to compare the relationship between them. Standard feed-forward neural networks are commonly used to model similarity between aligned (decomposed) sub-phrases and then aggregated into the final prediction layers.

Alignment between sentences has become a staple technique in NLI research and many recent state-of-the-art models such as the Enhanced Sequential Inference Model (ESIM) \cite{DBLP:conf/acl/ChenZLWJI17} also incorporate the alignment strategy. The difference here is that ESIM considers a non-parameterized comparison scheme, i.e., \textit{concatenating} the subtraction and element-wise product of aligned sub-phrases, along with two original sub-phrases, into the final comparison vector. A bidirectional LSTM is then used to aggregate the compared alignment vectors.

This paper presents a new neural model for NLI. There are several new novel components in our work. Firstly, we propose a \textit{compare, compress and propagate} (ComProp) architecture where compressed alignment features are propagated to upper layers (such as a RNN-based encoder) for enhancing representation learning. Secondly, in order to achieve an efficient propagation of alignment features, we propose alignment factorization layers to reduce each alignment vector to a single scalar valued feature. Each scalar valued feature is used to augment the base word representation, allowing the subsequent RNN encoder layers to benefit from not only global but also cross sentence information.

There are several major advantages to our proposed architecture. Firstly, our model is relatively compact, i.e., we compress alignment feature vectors and augment them to word representations instead. This is to avoid large alignment (or match) vectors being propagated across the network. As a result, our model is more parameter efficient compared to ESIM since the width of the middle layers of the network is now much smaller. To the best of our knowledge, this is the first work that explicitly employs such a paradigm.

Secondly, the explicit usage of compression enables improved interpretabilty since each alignment pair is compressed to a scalar and hence, can be easily visualised. Previous models such as ESIM use subtractive operations on alignment vectors, edging on the intuition that these vectors represent contradiction. Our model is capable of visually demonstrating this phenomena. As such, our design choice enables a new way of deriving insight from neural NLI models.

Thirdly, the alignment factorization layer is expressive and powerful, combining ideas from standard machine learning literature \cite{rendle2010factorization} with modern neural NLI models. The factorization layer tries to decompose the alignment vector (constructed from the variations of $a-b$, $a \odot b$ and $[a;b]$), learning higher-order feature interactions between each compared alignment. In other words, it models the second-order (pairwise) interactions between \textit{each} feature in \textit{every} alignment vector using factorized parameters, allowing more expressive comparison to be made over traditional fully-connected layers (FC). Moreover, factorization-based models are also known to be able to model low-rank structure and reduce risks of overfitting. The effectiveness of the factorization alignment over alternative baselines such as feed-forward neural networks is confirmed by early experiments.

The major contributions of this work are summarized as follows:
\begin{itemize}
\item We introduce a \textit{Compare, Compress and Propagate} (ComProp) architecture for NLI. The key idea is to use the myriad of generated comparison vectors for augmentation of the base word representation instead of simply aggregating them for prediction. Subsequently, a standard compositional encoder can then be used to learn representations from the augmented word representations. We show that we are able to derive meaningful insight from visualizing these augmented features.
\item For the first time, we adopt expressive factorization layers to model the relationships between soft-aligned sub-phrases of sentence pairs. Empirical experiments confirm the effectiveness of this new layer over standard fully connected layers.
\item Overall, we propose a new neural model - \textsc{CAFE} (\textbf{C}omProp \textbf{A}lignment-\textbf{F}actorized \textbf{E}ncoders) for NLI. Our model achieves state-of-the-art performance on SNLI, MultiNLI and the new SciTail dataset, outperforming existing state-of-the-art models such as ESIM. Ablation studies confirm the effectiveness of each proposed component in our model.
\end{itemize}

\section{Related Work}

Natural language inference (or textual entailment recognition) is a long standing problem in NLP research, typically carried out on smaller datasets using
traditional methods \cite{Maccartney:2009:NLI:1751277,Dagan:2005:PRT:2100045.2100054,MacCartney:2008:MSC:1599081.1599147,Iftene:2007:HTS:1654536.1654562}.

The relatively recent creation of $570K$ human annotated sentence pairs \cite{DBLP:conf/emnlp/BowmanAPM15} have spurred on many recent works that use neural networks for NLI. Many advanced neural architectures have been proposed for the NLI task, with most exploiting some variants of neural attention which learn to pay attention to important segments in a sentence \cite{DBLP:conf/emnlp/ParikhT0U16,DBLP:conf/acl/ChenZLWJI17,DBLP:conf/naacl/WangJ16,rocktaschel2015reasoning,DBLP:conf/eacl/YuM17a}.

Amongst the myriad of neural architectures proposed for NLI, the ESIM \cite{DBLP:conf/acl/ChenZLWJI17} model is one of the best performing models. The ESIM, primarily motivated by soft subphrase alignment in \cite{DBLP:conf/emnlp/ParikhT0U16}, learns alignments between BiLSTM encoded representations and aggregates them with another BiLSTM layer. The authors also propose the usage of subtractive composition, claiming that this helps model contradictions amongst alignments.

Compare-Aggregate models are also highly popular in NLI tasks. While this term was coined by \cite{DBLP:journals/corr/WangJ16b}, many prior NLI models follow this design \cite{DBLP:conf/ijcai/WangHF17,DBLP:conf/emnlp/ParikhT0U16,DBLP:journals/corr/abs-1709-04348,DBLP:conf/acl/ChenZLWJI17}. The key idea is to aggregate matching features and pass them through a dense layer for prediction. \cite{DBLP:conf/ijcai/WangHF17} proposed BiMPM, which adopts multi-perspective cosine matching across sequence pairs. \cite{DBLP:journals/corr/WangJ16b} proposed a one-way attention and convolutional aggregation layer. \cite{DBLP:journals/corr/abs-1709-04348} learns representations with highway layers and adopts ResNet for learning features over an interaction matrix.

There are several other notable models for NLI. For instance, models that leverage directional self-attention \cite{DBLP:journals/corr/abs-1709-04696} or Gumbel-Softmax \cite{choi2017unsupervised}. DGEM is a graph based attention model which was proposed together with a new entailment challenge dataset, SciTail \cite{scitail}. Pretraining have been known to also be highly useful in the NLI task. For instance, contextualized vectors learned from machine translation \cite{mccann2017learned} (CoVe) or language modeling \cite{peters2018deep} (ELMo) have showned to be able to improve performance when integrated with existing NLI models.

Our work compares and compresses alignment pairs using factorization layers which leverages the rich history of standard machine learning literature. Our factorization layers incorporate highly expressive factorization machines (FMs) \cite{rendle2010factorization} into neural NLI models. In standard machine learning tasks, FMs remain a very competitive choice for learning feature interactions \cite{xiao2017attentional} for both standard classification and regression problems. Intuitively, FMs are adept at handling data sparsity (typically interactions) by using factorized parameters to approximate a feature matching matrix. This makes it suitable in our model architecture since feature interaction between subphrase alignment pairs is typically very sparse as well.

A recent work \cite{beutel2018latent} reports an interesting empirical study pertaining to the ability of standard FC layers and their ability to model `cross features' (or multiplicative features). Their overall finding suggests that while standard ReLU FC layers are able to approximate 2-way or 3-way features, they are extremely inefficient in doing so (requiring either very wide or deep layers). This further motivates the usage of FMs in this work and is well aligned with our empirical results, i.e., strong competitive performance with reasonably small parameterization.

\section{Our Proposed Model}
\begin{figure}[ht]
  \centering
    \includegraphics[width=0.8\linewidth]{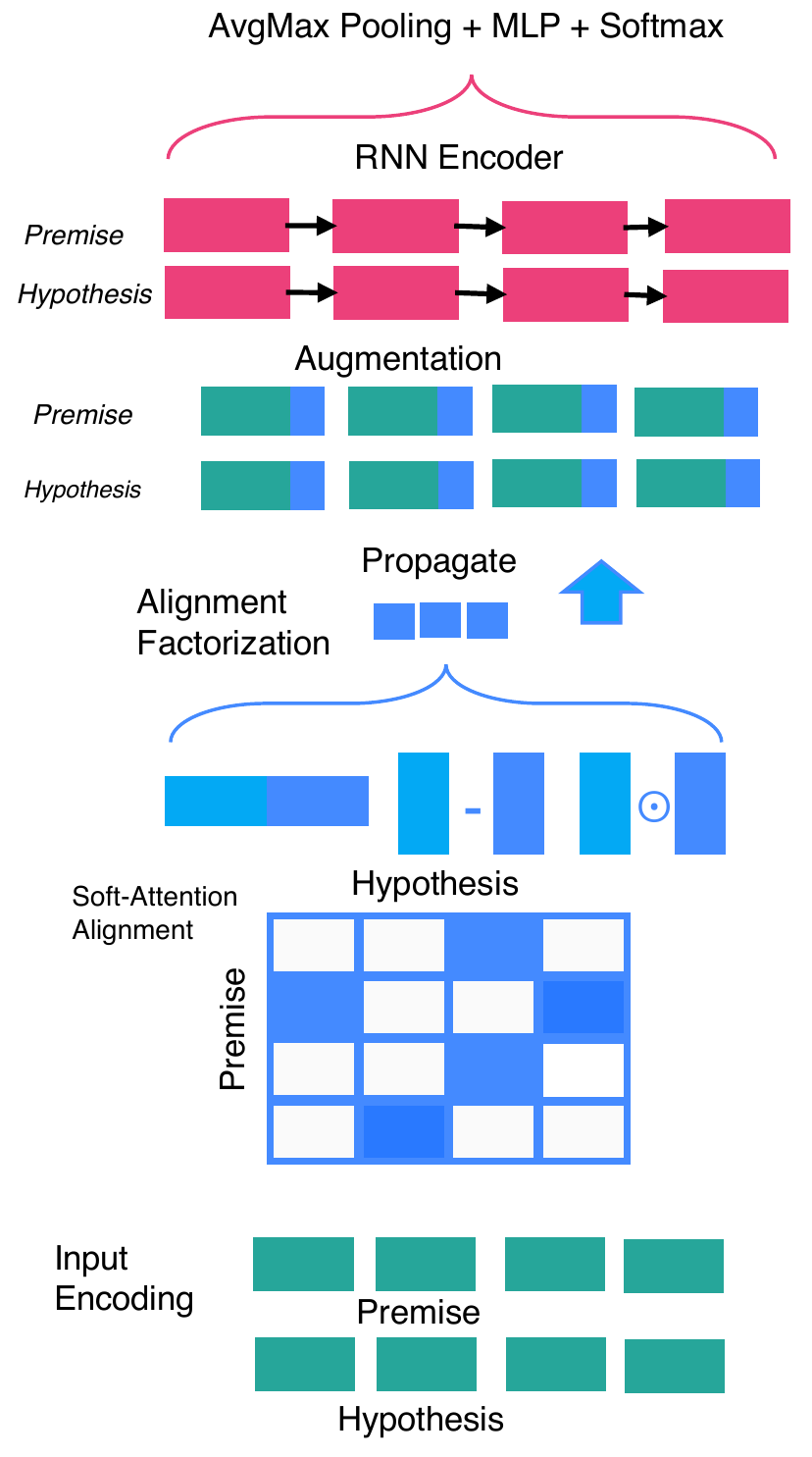}
    \caption{High level overview of our proposed architecture (\textit{best viewed in color}). Alignment vectors are compressed and then propagated to upper representation learning layers (RNN encoders). Intra-attention is omitted in this diagram due to the lack of space. }
    \label{fig:high_level}
\end{figure}

In this section, we provide a layer-by-layer description of our model architecture. Our model accepts two sentences as an input, i.e., $P$ (premise) and $H$ (hypothesis). Figure \ref{fig:high_level} illustrates a high-level overview of our proposed model architecture.

\subsection{Input Encoding Layer}{}
 This layer aims to learn a $k$-dimensional representation for each word. Following \cite{DBLP:journals/corr/abs-1709-04348}, we learn feature-rich word representations by concatenating word embeddings, character embeddings and syntactic (part-of-speech tag) embeddings (provided in the datasets). Character representations are learned using a convolutional encoder with max pooling function and is commonly used in many relevant literature \cite{DBLP:conf/ijcai/WangHF17,DBLP:conf/repeval/ChenZLWJI17}.
\paragraph{Highway Encoder}
 Subsequently, we pass each concatenated word vector into a two layer highway network \cite{DBLP:journals/corr/SrivastavaGS15} in order to learn a $k$-dimensional representation. Highway networks are gated projection layers which learn adaptively control how much information is being carried to the next layer. Our strategy is similar to \cite{DBLP:conf/emnlp/ParikhT0U16} which trains the projection layer in place of tuning the embedding matrix. The usage of highway layers over standard projection layers is empirically motivated. However, an intuition would be that the gates in this layer adapt to learn the relative importance of each word to the NLI task. Let $H(.)$ and $T(.)$ be single layered affine transforms with ReLU and sigmoid activation functions respectively. A single highway network layer is defined as:
 \begin{align}
y = H(x, W_{H}) \cdot T(x, W_{T}) + C \cdot x
 \end{align}
 where $C=(1-T(x, W_{T}))$ and $W_H, W_{T} \in \mathbb{R}^{r \times d}$ Notably, the dimensions of the affine transform might be different from the size of the input vector. In this case, an additional nonlinear transform is used to project $x$ to the same dimensionality. The output of this layer is $\bar{P} \in \mathbb{R}^{k \times \ell_{P}}$ (premise) and $\bar{H} \in \mathbb{R}^{k \times \ell_{H}}$ (hypothesis), with each word converted to a $r$-dimensional vector.

\subsection{Soft-Attention Alignment Layer}
This layer describes two soft-attention alignment techniques that are used in our model.
\paragraph{Inter-Attention Alignment Layer}
This layer learns an alignment of sub-phrases between $\bar{P}$ and $\bar{H}$. Let $F(.)$ be a standard projection layer with ReLU activation function. The alignment matrix of two sequences is defined as follows:
\begin{align}
e_{ij} &= F(\bar{p}_i)^\top\cdot F(\bar{h}_j)
\end{align}
where $E \in \mathbb{R}^{\ell_{p} \times \ell_h}$ and $\bar{p}_{i}$, $\bar{h}_j$ are the $i$-th and $j$-th word in the premise and hypothesis respectively.
\begin{align}
\beta_i &= \sum^{\ell_{p}}_{j=1} \frac{exp(e_{ij})}{\sum_{k=1}^{\ell_{p}} exp(e_{ik})} \bar{p}_{j} \\
\alpha_j &= \sum^{\ell_{h}}_{i=1} \frac{exp(e_{ij})}{\sum_{k=1}^{\ell_{h}} exp(e_{kj})} \bar{h}_{i}
\end{align}
where $\beta_i$ is the sub-phrase in $\bar{P}$ that is softly aligned to $h_i$.
Intuitively, $\beta_i$ is a weighted sum across $\{p_j\}^{\ell_{p}}_{j=1}$, selecting the most relevant parts of $\bar{P}$ to represent $h_i$.

\paragraph{Intra-Attention Alignment Layer}
This layer learns a \textit{self-alignment} of sentences and is applied to both $\bar{P}$ and $\bar{H}$ independently. For the sake of brevity, let $\bar{S}$ represent either $\bar{P}$ or $\bar{H}$, the intra-attention alignment is computed as:
\begin{align}
s^{\prime}_i &= \sum^{\ell_{p}}_{j=1} \frac{exp(f_{ij})}{\sum_{k=1}^{\ell_{p}} exp(f_{ik})} \bar{s}_{j}
\end{align}
where $f_{ij} = G(\bar{s}_i)^\top\cdot G(\bar{s}_j)$ and $G(.)$ is a nonlinear projection layer with ReLU activation function. The intra-attention layer models similarity of each word with respect to the entire sentence, capturing long distance dependencies and `global' context of the entire sentence.

\subsection{Alignment Factorization Layer}
This layer aims to learn a scalar valued feature for each comparison between aligned sub-phrases. Firstly, we introduce our factorization operation, which lives at the core of our neural model.
\paragraph{Factorization Operation}
Given an input vector $x$, the factorization operation \cite{rendle2010factorization} is defined as:
% \begin{align}
% L(x) &= w_{0} + \sum^{n}_{i=1} w_i \: x_i  \\
% P(x) & =  \sum^{n}_{i=1} \sum^{n}_{j=i+1} \langle v_i, v_j \rangle \: x_i \: x_j \\
% Z(x) &= L(x) + P(x)
% \end{align}
\begin{align}
Z(x) &= w_{0} + \sum^{n}_{i=1} w_i \: x_i  + \sum^{n}_{i=1} \sum^{n}_{j=i+1} \langle v_i, v_j \rangle \: x_i \:x_j
\end{align}
where $Z(x)$ is a scalar valued output, $\langle .;. \rangle$ is the dot product between two vectors and $w_{0}$ is the global bias. Factorization machines model low-rank structure within the matching vector producing a scalar feature.  The parameters of this layer are $w_{0} \in \mathbb{R}, w \in \mathbb{R}^{r}$ and $v \in \mathbb{R}^{r \times k}$. The first term $\sum^{n}_{i=1} w_i \: x_i$ is simply a linear term. The second term $ \sum^{n}_{i=1} \sum^{n}_{j=i+1} \langle v_i, v_j \rangle \: x_i \:x_j$ captures all pairwise interactions in $x$ (the input vector) using the factorization of matrix $v$.

\paragraph{Inter-Alignment Factorization}
This operation compares the alignment between inter-attention aligned representations, i.e., $(\beta_i, h_i)$ and $(\alpha_j, p_j)$. Let $(a, b)$ represent an alignment pair, we apply the following operations:
\begin{align}
y_{c}=Z([a;b]) \:;\:
y_{s}=Z(a - b) \:;\:
y_{m}=Z(a \odot b)
\end{align}
where $y_c, y_s, y_m \in \mathbb{R}$, $Z(.)$ is the factorization operation, $[.;.]$ is the concatenation operator and $\odot$ is the element-wise multiplication. The intuition of modeling subtraction is targeted at capturing contradiction. However, instead of simply concatenating the extra comparison vectors, we compress them using the factorization operation. Finally, for each alignment pair, we obtain three scalar-valued features which map precisely to a word in the sequence.
\paragraph{Intra-Alignment Factorization}
Next, for each sequence, we also apply alignment factorization on the intra-aligned sentences. Let $(s,s^\prime)$ represent an \textit{intra-aligned} pair from either the premise or hypothesis, we compute the following operations:
\begin{align}
v_{c}=Z([s;s^\prime]) \:;\:
v_{s}=Z(s - s^\prime) \:;\:
v_{m}=Z(s \odot s^\prime)
\end{align}
where $v_c, v_s, v_m \in \mathbb{R}$ and $Z(.)$ is the factorization operation. Applying alignment factorization to intra-aligned representations produces another three scalar-valued features which are mapped to each word in the sequence. Note that each of the \textit{six} factorization operations has its own parameters but shares them amongst all words in the sentences.

\subsection{Propagation and Augmentation}
Finally, the \textit{six} factorized features are then aggregated\footnote{Following \cite{DBLP:conf/emnlp/ParikhT0U16}, we may also concatenate the intra-aligned vector to $u_i$ which we found to have speed up convergence.} via concatenation to form a final feature vector that is propagated to upper representation learning layers via augmentation of the word representation $\bar{P}$ or $\bar{H}$.
\begin{align}
u_i = [s_i ; f^{i}_{intra}; f^{i}_{inter}]
\end{align}
where $s_i$ is $i$-th word in $\bar{P}$ or $\bar{H}$, and $f^{i}_{intra}$ and  $f^{i}_{inter}$ are the intra-aligned $[v_c; v_s; v_m]$  and inter-aligned $[y_c; y_s; y_m]$ features for the $i$-th word in the sequence respectively.
Intuitively, $f^i_{intra}$ augments each word with global knowledge of the sentence and $f^i_{inter}$ augments each word with cross-sentence knowledge via inter-attention.
\subsection{Sequential Encoder Layer}
For each sentence, the augmented word representations $u_1, u_2, \dots u_{\ell}$ are then passed into a sequential encoder layer. We adopt a standard vanilla LSTM encoder.
\begin{align}
h_i = LSTM(u, i), \forall i \in [1, \dots \ell]
\end{align}
where $\ell$ represents the maximum length of the sequence. Notably, the parameters of the LSTM are \textit{siamese} in nature, sharing weights between both premise and hypothesis. We do not use a bidirectional LSTM encoder, as we found that it did not lead to any improvements on the held-out set. A logical explanation would be because our word representations are already augmented with global (intra-attention) information. As such, modeling in the reverse direction is unnecessary, resulting in some computational savings.
\paragraph{Pooling Layer}
Next, to learn an overall representation of each sentence, we apply a pooling function across all hidden outputs of the sequential encoder. The pooling function is a concatenation of temporal max and average (avg) pooling.
\begin{align}
x = [\max([h_1, \cdots h_\ell]); avg([h_1, \cdots h_\ell])]
\end{align}
where $x$ is a final $2k$-dimensional representation of the sentence (premise or hypothesis). We also experimented with \textit{sum} and \textit{avg} standalone poolings and found \textit{sum} pooling to be relatively competitive.
\subsection{Prediction Layer}
Finally, given a fixed dimensional representation of the premise $x_p$ and hypothesis $x_h$, we pass their concatenation into a two-layer $h$-dimensional highway network. Since the highway network has been already defined earlier, we omit the technical details here. The final prediction layer of our model is computed as follows:
\begin{align}
y_{out} = H_{2}(H_{1}([x_{p}; x_{h}; x_p \odot x_h; x_p - x_h]))
\end{align}
where $H_{1}(.), H_{2}(.)$ are highway network layers with ReLU activation. The output is then passed into a final linear softmax layer.
\begin{align}
y_{pred} = softmax(W_{F} \cdot y_{out} + b_{F})
\end{align}
where $W_{F} \in \mathbb{R}^{h \times 3}$ and $b_{F} \in \mathbb{R}^{3}$. The network is then trained using standard multi-class cross entropy loss with L2 regularization.

\section{Experiments}

\begin{table*}[ht]
  \centering

    \begin{tabular}{|l|ccc|}
    \hline
    Model & Params & Train& Test\\
    \hline
    \multicolumn{4}{|c|}{\textbf{Single Model (w/o Cross Sentence Attention)}} \\
    \hline
    % 300D LSTM Encoder \cite{DBLP:conf/emnlp/BowmanAPM15}  & 3.0M  & 83.9  & 80.6 \\
    % 1024D GRU + skipthought \cite{DBLP:journals/corr/VendrovKFU15} & 15M   & 98.8  & 81.4 \\
    % 300D tree-based CNN \cite{DBLP:conf/acl/MouMLX0YJ16} & 3.5M  & 83.3  & 82.1 \\
    % 300D SPINN-PI encoders \cite{DBLP:conf/acl/BowmanGRGMP16}  & 3.7M  & 89.2  & 83.2 \\
    % 600D BiLSTM intra-attention \cite{DBLP:journals/corr/LiuSLW16}& 2.8M  & 84.5  & 84.2 \\
    % 300D NSE encoders  & 3.0M    & 86.2  & 84.6 \\
    % 600D Gated BiLSTM + intra-att \cite{DBLP:conf/repeval/ChenZLWJI17}  & 12M& 90.5& 85.5\\
    300D Gumbel TreeLSTM \cite{choi2017unsupervised} & 2.9M&91.2 &85.6 \\
    300D DISAN \cite{DBLP:journals/corr/abs-1709-04696}&2.4M &91.1 & 85.6\\
    300D Residual Stacked Encoders \cite{DBLP:conf/repeval/NieB17} & 9.7M & 89.8& 85.7\\
    600D Gumbel TreeLSTM \cite{choi2017unsupervised}& 10M & 93.1 & \textbf{86.0}\\
    300D \textsc{CAFE} (w/o CA)  & 3.7M& 87.3 & \underline{85.9}\\
    \hline
    \multicolumn{4}{|c|}{\textbf{Single Models}} \\
    \hline
    100D LSTM with attention \cite{rocktaschel2015reasoning} & 250K  & 85.3  & 83.5 \\
    % 50D Stacked DF-LSTM \cite{DBLP:conf/emnlp/LiuQZCH16} & 190K & 86.7 & 85.1 \\
    300D mLSTM  \cite{DBLP:conf/naacl/WangJ16}& 1.9M  & 92.0    & 86.1 \\
    450D LSTMN + deep att. fusion \cite{DBLP:conf/emnlp/0001DL16} & 3.4M  & 88.5  & 86.3 \\
    % 200D DecompAtt \cite{DBLP:conf/emnlp/ParikhT0U16} & 380K  & 89.5  & 86.3 \\
    200D DecompAtt + Intra-Att \cite{DBLP:conf/emnlp/ParikhT0U16}  & 580K  & 90.5  & 86.8 \\
    300D NTI-SLSTM-LSTM \cite{DBLP:conf/eacl/YuM17} & 3.2M  & 88.5  & 87.3 \\
    300D re-read LSTM \cite{DBLP:conf/coling/ShaCSL16} & 2.0M  & 90.7  & 87.5 \\
    BiMPM \cite{DBLP:conf/ijcai/WangHF17} & 1.6M & 90.9 & 87.5 \\
     448D DIIN \cite{DBLP:journals/corr/abs-1709-04348} & 4.4M  & 91.2  & 88.0 \\
    600D ESIM \cite{DBLP:conf/acl/ChenZLWJI17} & 4.3M  & 92.6  & 88.0 \\

    \hline
    150D \textsc{CAFE} (SUM+2x200D MLP)  & 750K & 88.2 & 87.7 \\
    200D \textsc{CAFE} (SUM+2x400D MLP)  & 1.4M & 89.4 & 88.1 \\
    300D \textsc{CAFE} (SUM+2x600D MLP)  & 3.5M & 89.2&  \underline{88.3}\\
    300D \textsc{CAFE} (AVGMAX+300D HN)  & 4.7M &89.8 &  \textbf{88.5}\\
    \hline
    \multicolumn{4}{|c|}{\textbf{Ensemble Models}} \\
    \hline
    600D ESIM + 300D Tree-LSTM \cite{DBLP:conf/acl/ChenZLWJI17}  & 7.7M & 93.5 & 88.6 \\
    BiMPM \cite{DBLP:conf/ijcai/WangHF17}  & 6.4M  & 93.2 & 88.8 \\
    448D DIIN \cite{DBLP:journals/corr/abs-1709-04348} & 17.0M   & 92.3 &  88.9 \\
    300D \textsc{CAFE} (Ensemble) & 17.5M   & 92.5 & \textbf{89.3} \\
    \hline
     \multicolumn{4}{|c|}{\textbf{External Resource Models}} \\
     \hline
      BiAttentive Classification + CoVe + Char \cite{mccann2017learned} & 22M & 88.5 & 88.1 \\
     KIM \cite{chen2017natural} & 4.3M& 94.1 & 88.6\\
     ESIM + ELMo \cite{peters2018deep} & 8.0M & 91.6 & 88.7 \\
         200D \textsc{CAFE} (AVGMAX + 200D MLP) + ELMo & 1.4M & 89.5 & \textbf{89.0} \\
    % KIM Ensemble \cite{chen2017natural}& 43M& 93.6 & 89.1\\
     \hline

    \end{tabular}%
     \caption{Performance comparison of all published models on the SNLI benchmark.}
  \label{tab:snli_results}%
\end{table*}%

In this section, we describe our experimental setup and report our experimental results.
% In this section, we describe our empirical evaluation.
\subsection{Experimental Setup}
To ascertain the effectiveness of our models, we use the SNLI \cite{DBLP:conf/emnlp/BowmanAPM15} and MultiNLI \cite{DBLP:journals/corr/WilliamsNB17} benchmarks which are standard and highly competitive benchmarks for the NLI task. We also include the newly released SciTail dataset \cite{scitail} which is a binary entailment classification task constructed from science questions. Notably, SciTail is known to be a difficult dataset for NLI, made evident by the low accuracy scores even though it is binary in nature.

\paragraph{SNLI} The state-of-the-art competitors on this dataset are the BiMPM \cite{DBLP:conf/ijcai/WangHF17}, ESIM \cite{DBLP:conf/acl/ChenZLWJI17} and DIIN \cite{DBLP:journals/corr/abs-1709-04348}. We compare against competitors across three settings. The first setting disallows cross sentence attention. In the second setting, cross sentence is allowed. The last (third) setting is a comparison between model ensembles while the first two settings only comprise single models. Note that we consider the 1st setting to be relatively less important (since our focus is not on the encoder itself) but still report the results for completeness.

\paragraph{MultiNLI} We compare on two test sets (\textit{matched} and \textit{mismatched}) which represent in-domain and out-domain performance. The main competitor on this dataset is the ESIM model, a powerful state-of-the-art SNLI baseline. We also compare with ESIM + Read \cite{DBLP:journals/corr/Weissenborn17}.
\paragraph{SciTail} This dataset only has one official setting. We compare against the reported results of ESIM \cite{DBLP:conf/acl/ChenZLWJI17} and DecompAtt \cite{DBLP:conf/emnlp/ParikhT0U16} in the original paper. We also compare with DGEM, the new model proposed in \cite{scitail}.

Across all experiments and in the spirit of fair comparison, we only compare with works that (1) do not use extra training data and (2) do not use external resources (such as external knowledge bases, etc.). However, for the sake of completeness, we still report their scores\footnote{Additionally, we added ELMo \cite{peters2018deep} to our CAFE model at the embedding layer. We report CAFE + ELMo under external resource models. This was done post review after EMNLP. Due to resource constraints, we did not train CAFE + ELMo ensembles but a single run (and single model) of CAFE + ELMo already achieves 89.0 score on SNLI. } \cite{mccann2017learned,chen2017natural,peters2018deep}.

\subsection{Implementation Details}
We implement our model in TensorFlow \cite{tensorflow2015-whitepaper} and train them on Nvidia P100 GPUs. We use the Adam optimizer \cite{DBLP:journals/corr/KingmaB14} with an initial learning rate of $0.0003$. L2 regularization is set to $10^{-6}$. Dropout with a keep probability of $0.8$ is applied after each fully-connected, recurrent or highway layer. The batch size is tuned amongst $\{128, 256, 512\}$. The number of latent factors $k$ for the factorization layer is tuned amongst $\{5,10,50,100,150\}$. The size of the hidden layers of the highway network layers are set to $300$. All parameters are initialized with xavier initialization.  Word embeddings are pre-loaded with $300d$ GloVe embeddings \cite{DBLP:conf/emnlp/PenningtonSM14} and fixed during training. Sequence lengths are padded to batch-wise maximum. The batch order is (randomly) sorted within buckets following \cite{DBLP:conf/emnlp/ParikhT0U16}.

\subsection{Experimental Results}
Table \ref{tab:snli_results} reports our results on the SNLI benchmark. On the cross sentence (single model setting), the performance of our proposed CAFE model is extremely competitive. We report the test accuracy of CAFE at different extents of parameterization, i.e., varying the size of the LSTM encoder, width of the pre-softmax hidden layers and final pooling layer. CAFE obtains $88.5\%$ accuracy on the SNLI test set, an extremely competitive score on the extremely popular benchmark. Notably, competitive results can be also achieved with a much smaller parameterization. For example, CAFE also achieves $88.3\%$ and $88.1\%$ test accuracy with only $3.5M$ and $1.5M$ parameters respectively. This outperforms the state-of-the-art ESIM and DIIN models with only a fraction of the parameter cost. At $88.1\%$, our model has about three times less parameters than ESIM/DIIN (i.e., 1.4M versus 4.3M/4.4M). Moreover, our lightweight adaptation achieves $87.7\%$ with only $750K$ parameters, which makes it extremely performant amongst models having the same amount of parameters such as the decomposable attention model ($86.8\%$).

% After removing the inter-attention layers from CAFE, the performance remains competitive to top performing encoder models. CAFE (w/o cross attention) achieves a respectable $85.9\%$ accuracy in this setting. Notably, the best performing model, the Gumbel TreeLSTM ($86.0\%)$ has over $10M$ parameters. On the other hand, our CAFE model has a $300\%$ less parameters, yet performs competitively to the Gumbel TreeLSTM.

Finally, an ensemble of $5$ CAFE models achieves $89.3\%$ test accuracy, the best test scores on the SNLI benchmark to date\footnote{As of 22nd May 2018, the deadline of the EMNLP submisssion.}. Overall, we believe that the good performance of our CAFE can be attributed to (1) the effectiveness of the ComProp architecture (i.e., providing word representations with global and local knowledge for better representation learning) and (2) the expressiveness of alignment factorization layers that are used to decompose and compare word alignments. More details are given at the ablation study. Finally, we emphasize that CAFE is also relatively lightweight, efficient and fast to train given its performance. A single run on SNLI takes approximately $5$ minutes per epoch with a batch size of $256$. Overall, a single run takes $\approx 3$ hours to get to convergence.

\begin{table}[htbp]
  \centering
\small
    \begin{tabular}{|l|cc|c|}
    \hline
    & \multicolumn{2}{c}{MultiNLI} & SciTail \\
    \hline
    Model & \multicolumn{1}{l}{Match} & \multicolumn{1}{|l|}{Mismatch} & - \\
    \hline
    Majority & 36.5 & 35.6 & 60.3 \\
    NGRAM$^\#$ & - & - & 70.6 \\
    % & & & \\
    CBOW$^{\flat}$ & 65.2 & 64.8 & - \\
    BiLSTM$^{\flat}$  & 69.8 & 69.4 & -\\
    \hline
     ESIM$^{\#,\flat}$ & 72.4  & 72.1 & 70.6 \\
    DecompAtt$^\#$ - & - & - &72.3\\
    DGEM$^\#$ & - & - &70.8 \\
    DGEM + Edge$^\#$ & -  & - &77.3 \\
   \hline
    ESIM$^\dagger$  & 76.3  & 75.8 & -\\
    ESIM + Read$^\dagger$  & 77.8 & 77.0 & - \\
    \hline
    \textsc{CAFE} & 78.7  & 77.9 & \textbf{83.3} \\
    CAFE Ensemble & \textbf{80.2}& \textbf{79.0} & - \\
    \hline
    \end{tabular}%
  \caption{Performance comparison (accuracy) on MultiNLI and SciTail. Models with $\dagger$, $\#$ and $\flat$ are reported from \cite{DBLP:journals/corr/Weissenborn17}, \cite{scitail} and \cite{DBLP:journals/corr/WilliamsNB17} respectively.}%
  \label{multinli_results}
\end{table}%

Table \ref{multinli_results} reports our results on the MultiNLI and SciTail datasets. On MultiNLI, CAFE significantly outperforms ESIM, a strong state-of-the-art model on both settings. We also outperform the ESIM + Read model \cite{DBLP:journals/corr/Weissenborn17}. An ensemble of CAFE models achieve competitive result on the MultiNLI dataset. On SciTail, our proposed CAFE model achieves state-of-the-art performance. The performance gain over strong baselines such as DecompAtt and ESIM are $\approx 10\%-13\%$ in terms of accuracy. CAFE also outperforms DGEM, which uses a graph-based attention for improved performance, by a significant margin of $5\%$. As such, empirical results demonstrate the effectiveness of our proposed CAFE model on the challenging SciTail dataset.


\subsection{Ablation Study}
\begin{table}[htbp]
  \centering
  \small

    \begin{tabular}{|p{4cm}|cc|}
    \hline
          & \multicolumn{1}{l}{Match} & \multicolumn{1}{l|}{Mismatch} \\
          \hline
    Original Model & 79.0    & 78.9 \\
    \hline
    (1a) Rm FM for 1L-FC & 77.7  & 77.9 \\
    % (1b) Rm FM for 2L-FC & 76.2  & 76.3 \\
    (1b) Rm FM for 1L-FC (ReLU) & 77.3  & 77.5 \\
    (1c) Rm FM for 2L-FC (ReLU) & 76.6  & 76.4 \\
    \hline
    (2) Remove Char Embed & 78.1  & 78.3 \\
        (3) Remove Syn Embed & 78.3  & 78.4 \\
    (4) Remove Inter Att & 75.2  & 75.6 \\
    % Remove Intra Att &       &  \\
    (5) Replace HW Pred. with FC & 77.7  & 77.9 \\
    (6) Replace HW Enc. with FC & 78.7  & 78.7 \\
    (7) Remove Sub Feat & 77.9  & 78.3 \\
    (8) Remove Mul Feat & 78.7  & 78.6 \\
    (9) Remove Concat Feat & 77.9  & 77.6 \\
    (10) Add Bi-directional & 78.3  & 78.4 \\
    \hline
    \end{tabular}%
      \caption{Ablation study on MultiNLI development sets. HW stands for Highway.}
  \label{tab:ablation}%
\end{table}%
Table \ref{tab:ablation} reports ablation studies on the MultiNLI development sets. In (1), we replaced all FM functions with regular full-connected (FC) layers in order to observe the effect of \textbf{FM versus FC}. More specifically, we experimented with several FC configurations as follows: (a) 1-layer linear, (b) 1-layer ReLU (c) 2-layer ReLU. The 1-layer linear setting performs the best and is therefore reported in Table \ref{tab:ablation}. Using ReLU seems to be worse than nonlinear FC layers. Overall, the best combination (option a) still experienced a decline in performance in both development sets.

In (2-3), we explore the utility of using character and syntactic embeddings, which we found to have helped CAFE marginally. In (4), we remove the inter-attention alignment features, which naturally impact the model performance significantly. In (5-6), we explore the effectiveness of the highway layers (in prediction layers and encoding layers) by replacing them to FC layers. We observe that both highway layers have marginally helped the overall performance. Finally, in (7-9), we remove the alignment features based on their composition type. We observe that the \textit{Sub} and \textit{Concat} compositions were more important than the \textit{Mul} composition. However, removing any of the three will result in some performance degradation. Finally, in (10), we replace the LSTM encoder with a BiLSTM, observing that adding bi-directionality did not improve performance for our model.

\subsection{Linguistic Error Analysis}


We perform a linguistic error analysis using the supplementary annotations provided by the MultiNLI dataset. We compare against the model outputs of the ESIM model across 13 categories of linguistic phenenoma \cite{DBLP:journals/corr/WilliamsNB17}.
Table \ref{tab:error} reports the result of our error analysis. We observe that our CAFE model generally outperforms ESIM on \textit{most categories}.

\begin{table}[htbp]
  \centering
  \small

    \begin{tabular}{|l|cccc|}
    \hline
          & \multicolumn{2}{c}{Matched}        & \multicolumn{2}{c|}{Mismatched}  \\

          & \multicolumn{1}{c}{ESIM} & \multicolumn{1}{c}{CAFE} & \multicolumn{1}{c}{ESIM} & \multicolumn{1}{c|}{CAFE} \\
          \hline
    Conditional & 100   & 70    & 60    & \textbf{85} \\
    Word overlap & 50    & \textbf{82} & 62    & \textbf{87} \\
    Negation & \textbf{76}    & \textbf{76}    & 71    & \textbf{80} \\
    Antonym & 67    & \textbf{82} & 58    & \textbf{80} \\
    Long Sentence & 75    & \textbf{79} & 69    & \textbf{77} \\
    Tense Difference & 73    & \textbf{82} & 79    & \textbf{89} \\
    Active/Passive & 88    & \textbf{100} & \textbf{91}    & 90 \\
    Paraphrase & \textbf{89}    & 88    & 84    & \textbf{95} \\
    Quantity/Time  & 33    & \textbf{53} & 54    & \textbf{62} \\
    Coreference & \textbf{83}    & 80    & 75    & \textbf{83} \\
    Quantifier & 69    & \textbf{75} & 72    & \textbf{80} \\
    Modal & 78    & \textbf{81} & 76    & \textbf{81} \\
    Belief & 65    & \textbf{77} & 67    & \textbf{83} \\
    \hline
    \end{tabular}%
     \caption{Linguistic Error Analysis on MultiNLI dataset. }
  \label{tab:error}%
\end{table}%

On the mismatched setting, CAFE outperforms ESIM in 12 out of 13 categories, losing only in one percentage point in \textit{Active/Passive} category. On the matched setting, CAFE is outperformed by ESIM very marginally on coreference and paraphrase categories. Despite generally achieving much superior results, we noticed that CAFE performs poorly on \textit{conditionals\footnote{This only accounts for $5\%$ of samples.}} on the matched setting. Measuring the absolute ability of CAFE, we find that CAFE performs extremely well in handling linguistic patterns of \textit{paraphrase detection} and \textit{active/passive}. This is likely to be attributed by the alignment strategy that CAFE and ESIM both exploits.

\subsection{Interpreting and Visualizing with CAFE}

Finally, we also observed that the propagated features are highly interpretable, giving insights to the inner workings of the CAFE model.
\begin{figure}[ht!]
  \centering
 \begin{subfigure}[b]{0.44\textwidth}
   \includegraphics[width=1\linewidth, height=0.6\textwidth]{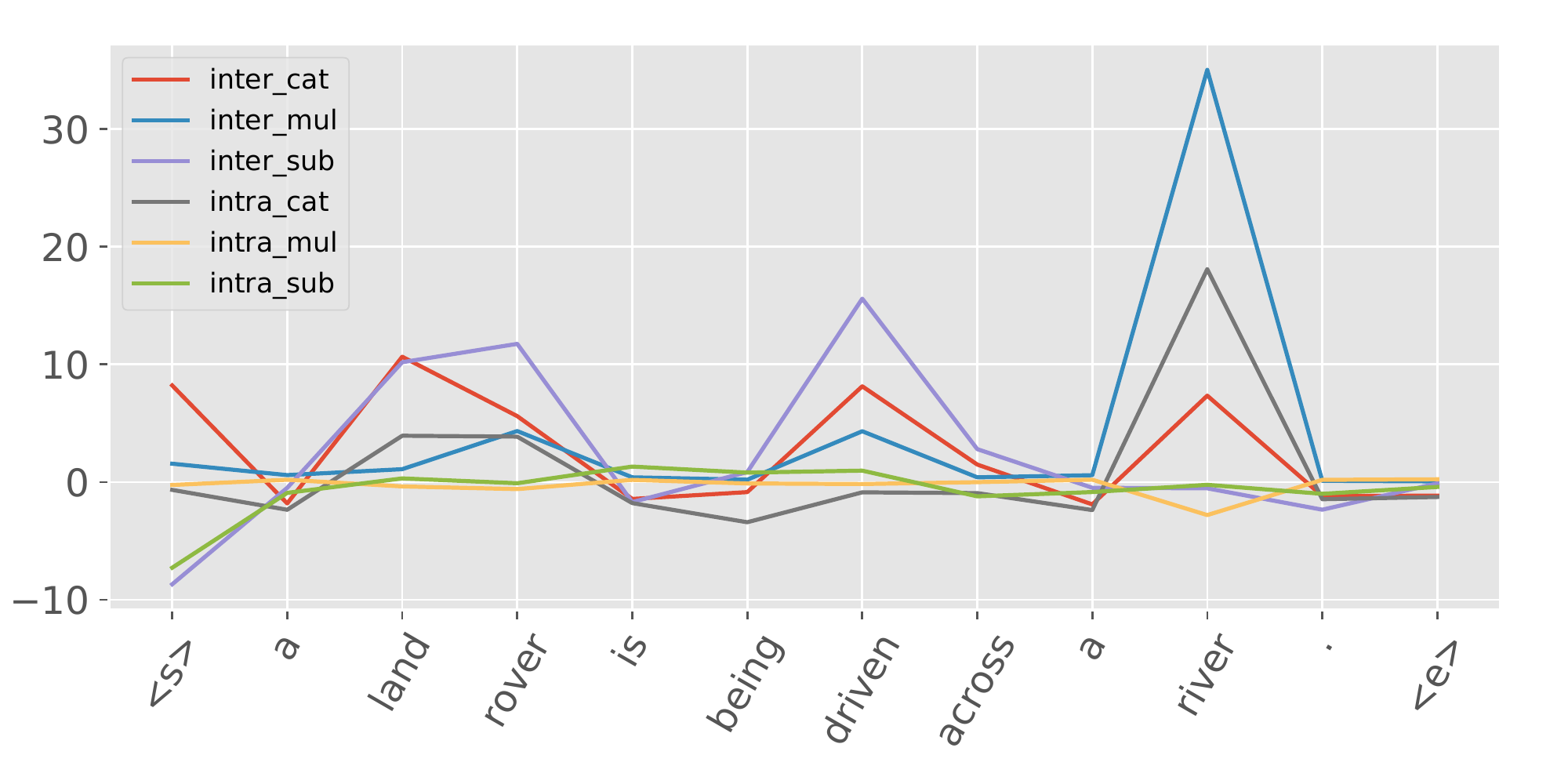}
   \label{fig:ex1a}
\end{subfigure}
\vspace{-1em}
\begin{subfigure}[b]{0.44\textwidth}
   \includegraphics[width=1\linewidth,height=0.6\textwidth]{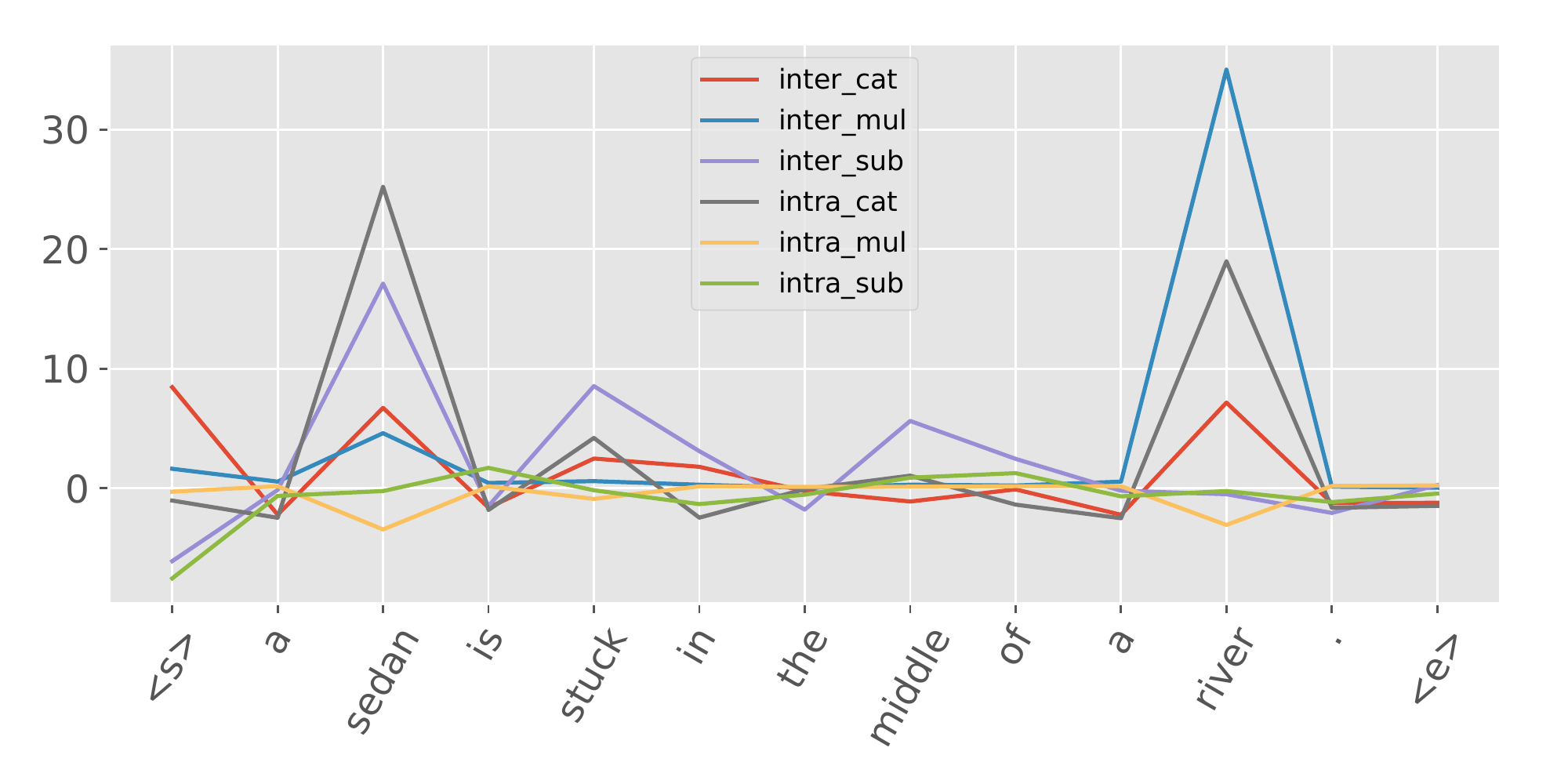}
   \label{fig:ex1bb}
\end{subfigure}
\caption{Visualization of \textit{six} Propgated Features (\textit{Best viewed in color}). Legend is denoted by $\{$inter,intra$\}$ followed by the operations \textit{mul}, \textit{sub} or \textit{cat} (concat).} %\vspace{-3mm}
\label{cafe_analysis}
\end{figure}
Figure \ref{cafe_analysis} shows a visualization of the feature values from an example in the SNLI test set. The ground truth is \textit{contradiction}. Based on the above example we make several observations. Firstly, \textit{inter\_mul} features mostly capture identical words (or semantically similar words), i.e., \textit{inter\_mul} features for \textit{`river'} spikes in both sentences. Secondly, \textit{inter\_sub} spikes on conflicting words that might cause contradiction, e.g., \textit{`sedan'} and \textit{`land rover'} are not the same vehicle. Another interesting observation is that we notice the \textit{inter\_sub} features for \textit{driven} and \textit{stuck} spiking. This also validates the observation of \cite{DBLP:conf/acl/ChenZLWJI17}, which shows what the \textit{sub} vector in the ESIM model is looking out for contradictory information. However, our architecture allows the inspection of these vectors since they are compressed via factorization, leading to larger extents of explainability - a quality that neural models inherently lack. We also observed that intra-attention (e.g., intra\_cat) features seem to capture the more important words in the sentence (\textit{`river'}, \textit{`sedan'}, \textit{`land rover'}).

\section{Conclusion}
We proposed a new neural architecture, CAFE for NLI. CAFE achieves very competitive performance on three
benchmark datasets. Extensive ablation studies confirm the effectiveness of FM layers over FC layers. Qualitatively, we show how different compositional operators (e.g., sub and mul) behave in NLI task and shed light on why subtractive composition helps in other models such as ESIM.

\bibliographystyle{acl_natbib}
\bibliography{acl2018}

\begin{thebibliography}{}
\expandafter\ifx\csname natexlab\endcsname\relax\def\natexlab#1{#1}\fi

\bibitem[{Abadi et~al.(2015)Abadi, Agarwal, Barham, Brevdo, Chen, Citro,
  Corrado, Davis, Dean, Devin, Ghemawat, Goodfellow, Harp, Irving, Isard, Jia,
  Jozefowicz, Kaiser, Kudlur, Levenberg, Man\'{e}, Monga, Moore, Murray, Olah,
  Schuster, Shlens, Steiner, Sutskever, Talwar, Tucker, Vanhoucke, Vasudevan,
  Vi\'{e}gas, Vinyals, Warden, Wattenberg, Wicke, Yu, and
  Zheng}]{tensorflow2015-whitepaper}
Mart\'{\i}n Abadi, Ashish Agarwal, Paul Barham, Eugene Brevdo, Zhifeng Chen,
  Craig Citro, Greg~S. Corrado, Andy Davis, Jeffrey Dean, Matthieu Devin,
  Sanjay Ghemawat, Ian Goodfellow, Andrew Harp, Geoffrey Irving, Michael Isard,
  Yangqing Jia, Rafal Jozefowicz, Lukasz Kaiser, Manjunath Kudlur, Josh
  Levenberg, Dan Man\'{e}, Rajat Monga, Sherry Moore, Derek Murray, Chris Olah,
  Mike Schuster, Jonathon Shlens, Benoit Steiner, Ilya Sutskever, Kunal Talwar,
  Paul Tucker, Vincent Vanhoucke, Vijay Vasudevan, Fernanda Vi\'{e}gas, Oriol
  Vinyals, Pete Warden, Martin Wattenberg, Martin Wicke, Yuan Yu, and Xiaoqiang
  Zheng. 2015.
\newblock {TensorFlow}: Large-scale machine learning on heterogeneous systems.
\newblock Software available from tensorflow.org.

\bibitem[{Beutel et~al.(2018)Beutel, Covington, Jain, Xu, Li, Gatto, and
  Chi}]{beutel2018latent}
Alex Beutel, Paul Covington, Sagar Jain, Can Xu, Jia Li, Vince Gatto, and
  H~Chi. 2018.
\newblock Latent cross: Making use of context in recurrent recommender systems
  .

\bibitem[{Bowman et~al.(2015)Bowman, Angeli, Potts, and
  Manning}]{DBLP:conf/emnlp/BowmanAPM15}
Samuel~R. Bowman, Gabor Angeli, Christopher Potts, and Christopher~D. Manning.
  2015.
\newblock A large annotated corpus for learning natural language inference.
\newblock In {\em Proceedings of the 2015 Conference on Empirical Methods in
  Natural Language Processing, {EMNLP} 2015, Lisbon, Portugal, September 17-21,
  2015\/}. pages 632--642.

\bibitem[{Chen et~al.(2017{\natexlab{a}})Chen, Zhu, Ling, and
  Inkpen}]{chen2017natural}
Qian Chen, Xiaodan Zhu, Zhen-Hua Ling, and Diana Inkpen. 2017{\natexlab{a}}.
\newblock Natural language inference with external knowledge.
\newblock {\em arXiv preprint arXiv:1711.04289\/} .

\bibitem[{Chen et~al.(2017{\natexlab{b}})Chen, Zhu, Ling, Wei, Jiang, and
  Inkpen}]{DBLP:conf/acl/ChenZLWJI17}
Qian Chen, Xiaodan Zhu, Zhen{-}Hua Ling, Si~Wei, Hui Jiang, and Diana Inkpen.
  2017{\natexlab{b}}.
\newblock \href{https://doi.org/10.18653/v1/P17-1152}{Enhanced {LSTM} for
  natural language inference}.
\newblock In {\em Proceedings of the 55th Annual Meeting of the Association for
  Computational Linguistics, {ACL} 2017, Vancouver, Canada, July 30 - August 4,
  Volume 1: Long Papers\/}. pages 1657--1668.
\newblock
  \href{https://doi.org/10.18653/v1/P17-1152}{https://doi.org/10.18653/v1/P17-1152}.

\bibitem[{Chen et~al.(2017{\natexlab{c}})Chen, Zhu, Ling, Wei, Jiang, and
  Inkpen}]{DBLP:conf/repeval/ChenZLWJI17}
Qian Chen, Xiaodan Zhu, Zhen{-}Hua Ling, Si~Wei, Hui Jiang, and Diana Inkpen.
  2017{\natexlab{c}}.
\newblock Recurrent neural network-based sentence encoder with gated attention
  for natural language inference.
\newblock In {\em Proceedings of the 2nd Workshop on Evaluating Vector Space
  Representations for NLP, RepEval@EMNLP 2017, Copenhagen, Denmark, September
  8, 2017\/}. pages 36--40.

\bibitem[{Cheng et~al.(2016)Cheng, Dong, and Lapata}]{DBLP:conf/emnlp/0001DL16}
Jianpeng Cheng, Li~Dong, and Mirella Lapata. 2016.
\newblock Long short-term memory-networks for machine reading.
\newblock In {\em Proceedings of the 2016 Conference on Empirical Methods in
  Natural Language Processing, {EMNLP} 2016, Austin, Texas, USA, November 1-4,
  2016\/}. pages 551--561.

\bibitem[{Choi et~al.(2017)Choi, Yoo, and Lee}]{choi2017unsupervised}
Jihun Choi, Kang~Min Yoo, and Sang-goo Lee. 2017.
\newblock Unsupervised learning of task-specific tree structures with
  tree-lstms.
\newblock {\em arXiv preprint arXiv:1707.02786\/} .

\bibitem[{Dagan et~al.(2006)Dagan, Glickman, and
  Magnini}]{Dagan:2005:PRT:2100045.2100054}
Ido Dagan, Oren Glickman, and Bernardo Magnini. 2006.
\newblock The pascal recognising textual entailment challenge.
\newblock In {\em Proceedings of the First International Conference on Machine
  Learning Challenges: Evaluating Predictive Uncertainty Visual Object
  Classification, and Recognizing Textual Entailment\/}. Springer-Verlag,
  Berlin, Heidelberg, MLCW'05, pages 177--190.

\bibitem[{Gong et~al.(2017)Gong, Luo, and
  Zhang}]{DBLP:journals/corr/abs-1709-04348}
Yichen Gong, Heng Luo, and Jian Zhang. 2017.
\newblock Natural language inference over interaction space.
\newblock {\em CoRR\/} abs/1709.04348.

\bibitem[{Iftene and Balahur-Dobrescu(2007)}]{Iftene:2007:HTS:1654536.1654562}
Adrian Iftene and Alexandra Balahur-Dobrescu. 2007.
\newblock Hypothesis transformation and semantic variability rules used in
  recognizing textual entailment.
\newblock In {\em Proceedings of the ACL-PASCAL Workshop on Textual Entailment
  and Paraphrasing\/}. Association for Computational Linguistics, Stroudsburg,
  PA, USA, RTE '07, pages 125--130.

\bibitem[{Khot et~al.(2018)Khot, Sabharwal, and Clark}]{scitail}
Tushar Khot, Ashish Sabharwal, and Peter Clark. 2018.
\newblock Scitail: A textual entailment dataset from science question
  answering.
\newblock In {\em AAAI\/}.

\bibitem[{Kingma and Ba(2014)}]{DBLP:journals/corr/KingmaB14}
Diederik~P. Kingma and Jimmy Ba. 2014.
\newblock Adam: {A} method for stochastic optimization.
\newblock {\em CoRR\/} abs/1412.6980.

\bibitem[{Maccartney(2009)}]{Maccartney:2009:NLI:1751277}
Bill Maccartney. 2009.
\newblock {\em Natural Language Inference\/}.
\newblock Ph.D. thesis, Stanford, CA, USA.
\newblock AAI3364139.

\bibitem[{MacCartney and Manning(2008)}]{MacCartney:2008:MSC:1599081.1599147}
Bill MacCartney and Christopher~D. Manning. 2008.
\newblock Modeling semantic containment and exclusion in natural language
  inference.
\newblock In {\em Proceedings of the 22Nd International Conference on
  Computational Linguistics - Volume 1\/}. Association for Computational
  Linguistics, Stroudsburg, PA, USA, COLING '08, pages 521--528.

\bibitem[{McCann et~al.(2017)McCann, Bradbury, Xiong, and
  Socher}]{mccann2017learned}
Bryan McCann, James Bradbury, Caiming Xiong, and Richard Socher. 2017.
\newblock Learned in translation: Contextualized word vectors.
\newblock In {\em Advances in Neural Information Processing Systems\/}. pages
  6297--6308.

\bibitem[{Nie and Bansal(2017)}]{DBLP:conf/repeval/NieB17}
Yixin Nie and Mohit Bansal. 2017.
\newblock Shortcut-stacked sentence encoders for multi-domain inference.
\newblock In {\em Proceedings of the 2nd Workshop on Evaluating Vector Space
  Representations for NLP, RepEval@EMNLP 2017, Copenhagen, Denmark, September
  8, 2017\/}. pages 41--45.

\bibitem[{Parikh et~al.(2016)Parikh, T{\"{a}}ckstr{\"{o}}m, Das, and
  Uszkoreit}]{DBLP:conf/emnlp/ParikhT0U16}
Ankur~P. Parikh, Oscar T{\"{a}}ckstr{\"{o}}m, Dipanjan Das, and Jakob
  Uszkoreit. 2016.
\newblock A decomposable attention model for natural language inference.
\newblock In {\em Proceedings of the 2016 Conference on Empirical Methods in
  Natural Language Processing, {EMNLP} 2016, Austin, Texas, USA, November 1-4,
  2016\/}. pages 2249--2255.

\bibitem[{Pennington et~al.(2014)Pennington, Socher, and
  Manning}]{DBLP:conf/emnlp/PenningtonSM14}
Jeffrey Pennington, Richard Socher, and Christopher~D. Manning. 2014.
\newblock Glove: Global vectors for word representation.
\newblock In {\em Proceedings of the 2014 Conference on Empirical Methods in
  Natural Language Processing, {EMNLP} 2014, October 25-29, 2014, Doha, Qatar,
  {A} meeting of SIGDAT, a Special Interest Group of the {ACL}\/}. pages
  1532--1543.

\bibitem[{Peters et~al.(2018)Peters, Neumann, Iyyer, Gardner, Clark, Lee, and
  Zettlemoyer}]{peters2018deep}
Matthew~E Peters, Mark Neumann, Mohit Iyyer, Matt Gardner, Christopher Clark,
  Kenton Lee, and Luke Zettlemoyer. 2018.
\newblock Deep contextualized word representations.
\newblock {\em arXiv preprint arXiv:1802.05365\/} .

\bibitem[{Rendle(2010)}]{rendle2010factorization}
Steffen Rendle. 2010.
\newblock Factorization machines.
\newblock In {\em Data Mining (ICDM), 2010 IEEE 10th International Conference
  on\/}. IEEE, pages 995--1000.

\bibitem[{Rockt{\"a}schel et~al.(2015)Rockt{\"a}schel, Grefenstette, Hermann,
  Ko{\v{c}}isk{\`y}, and Blunsom}]{rocktaschel2015reasoning}
Tim Rockt{\"a}schel, Edward Grefenstette, Karl~Moritz Hermann, Tom{\'a}{\v{s}}
  Ko{\v{c}}isk{\`y}, and Phil Blunsom. 2015.
\newblock Reasoning about entailment with neural attention.
\newblock {\em arXiv preprint arXiv:1509.06664\/} .

\bibitem[{Sha et~al.(2016)Sha, Chang, Sui, and Li}]{DBLP:conf/coling/ShaCSL16}
Lei Sha, Baobao Chang, Zhifang Sui, and Sujian Li. 2016.
\newblock Reading and thinking: Re-read {LSTM} unit for textual entailment
  recognition.
\newblock In {\em {COLING} 2016, 26th International Conference on Computational
  Linguistics, Proceedings of the Conference: Technical Papers, December 11-16,
  2016, Osaka, Japan\/}. pages 2870--2879.

\bibitem[{Shen et~al.(2017)Shen, Zhou, Long, Jiang, Pan, and
  Zhang}]{DBLP:journals/corr/abs-1709-04696}
Tao Shen, Tianyi Zhou, Guodong Long, Jing Jiang, Shirui Pan, and Chengqi Zhang.
  2017.
\newblock Disan: Directional self-attention network for rnn/cnn-free language
  understanding.
\newblock {\em CoRR\/} abs/1709.04696.

\bibitem[{Srivastava et~al.(2015)Srivastava, Greff, and
  Schmidhuber}]{DBLP:journals/corr/SrivastavaGS15}
Rupesh~Kumar Srivastava, Klaus Greff, and J{\"{u}}rgen Schmidhuber. 2015.
\newblock Highway networks.
\newblock {\em CoRR\/} abs/1505.00387.

\bibitem[{Wang and Jiang(2016{\natexlab{a}})}]{DBLP:journals/corr/WangJ16b}
Shuohang Wang and Jing Jiang. 2016{\natexlab{a}}.
\newblock A compare-aggregate model for matching text sequences.
\newblock {\em CoRR\/} abs/1611.01747.

\bibitem[{Wang and Jiang(2016{\natexlab{b}})}]{DBLP:conf/naacl/WangJ16}
Shuohang Wang and Jing Jiang. 2016{\natexlab{b}}.
\newblock Learning natural language inference with {LSTM}.
\newblock In {\em {NAACL} {HLT} 2016, The 2016 Conference of the North American
  Chapter of the Association for Computational Linguistics: Human Language
  Technologies, San Diego California, USA, June 12-17, 2016\/}. pages
  1442--1451.

\bibitem[{Wang et~al.(2017)Wang, Hamza, and Florian}]{DBLP:conf/ijcai/WangHF17}
Zhiguo Wang, Wael Hamza, and Radu Florian. 2017.
\newblock \href{https://doi.org/10.24963/ijcai.2017/579}{Bilateral
  multi-perspective matching for natural language sentences}.
\newblock In {\em Proceedings of the Twenty-Sixth International Joint
  Conference on Artificial Intelligence, {IJCAI} 2017, Melbourne, Australia,
  August 19-25, 2017\/}. pages 4144--4150.
\newblock
  \href{https://doi.org/10.24963/ijcai.2017/579}{https://doi.org/10.24963/ijcai.2017/579}.

\bibitem[{Weissenborn(2017)}]{DBLP:journals/corr/Weissenborn17}
Dirk Weissenborn. 2017.
\newblock Reading twice for natural language understanding.
\newblock {\em CoRR\/} abs/1706.02596.

\bibitem[{Williams et~al.(2017)Williams, Nangia, and
  Bowman}]{DBLP:journals/corr/WilliamsNB17}
Adina Williams, Nikita Nangia, and Samuel~R. Bowman. 2017.
\newblock A broad-coverage challenge corpus for sentence understanding through
  inference.
\newblock {\em CoRR\/} abs/1704.05426.

\bibitem[{Xiao et~al.(2017)Xiao, Ye, He, Zhang, Wu, and
  Chua}]{xiao2017attentional}
Jun Xiao, Hao Ye, Xiangnan He, Hanwang Zhang, Fei Wu, and Tat-Seng Chua. 2017.
\newblock Attentional factorization machines: Learning the weight of feature
  interactions via attention networks.
\newblock {\em arXiv preprint arXiv:1708.04617\/} .

\bibitem[{Yu and Munkhdalai(2017{\natexlab{a}})}]{DBLP:conf/eacl/YuM17a}
Hong Yu and Tsendsuren Munkhdalai. 2017{\natexlab{a}}.
\newblock Neural semantic encoders.
\newblock In {\em Proceedings of the 15th Conference of the European Chapter of
  the Association for Computational Linguistics, {EACL} 2017, Valencia, Spain,
  April 3-7, 2017, Volume 1: Long Papers\/}. pages 397--407.

\bibitem[{Yu and Munkhdalai(2017{\natexlab{b}})}]{DBLP:conf/eacl/YuM17}
Hong Yu and Tsendsuren Munkhdalai. 2017{\natexlab{b}}.
\newblock Neural tree indexers for text understanding.
\newblock In {\em Proceedings of the 15th Conference of the European Chapter of
  the Association for Computational Linguistics, {EACL} 2017, Valencia, Spain,
  April 3-7, 2017, Volume 1: Long Papers\/}. pages 11--21.

\end{thebibliography}

\end{document}